\pdfoutput=1

\documentclass[11pt]{article}

\usepackage[final]{acl}

\usepackage{times}
\usepackage{latexsym}
\usepackage{tabularx}

\usepackage[T1]{fontenc}

\usepackage[utf8]{inputenc}

\usepackage{microtype}

\usepackage{inconsolata}

\usepackage{graphicx}

\usepackage{nicefrac}
\usepackage{float}
\usepackage{subcaption}
\usepackage{multirow}
\usepackage{adjustbox}
\usepackage[linesnumbered,ruled,vlined]{algorithm2e}
\usepackage{soul}
\usepackage{marvosym}
\usepackage{amsfonts}
\usepackage{amsmath}
\usepackage{xcolor}
\usepackage{bbm}
\usepackage{siunitx}
\usepackage{lipsum}  

\DeclareMathOperator*{\argmin}{argmin} 

\title{LLM-TOPLA: Efficient LLM Ensemble by Maximising Diversity}

\author{Selim Furkan Tekin, Fatih Ilhan, Tiansheng Huang,  Sihao Hu, Ling Liu \\
        Georgia Institute of Technology, USA \\
    \small{
   \url{{stekin6, filhan3, thuang374, shu335, ll72}@gatech.edu}
 }
        }

\begin{document}
\maketitle
\begin{abstract}
Combining large language models during training or at inference time has shown substantial performance gain over component LLMs. This paper presents LLM-TOPLA, a diversity-optimized LLM ensemble method with three unique properties: (i) We introduce the focal diversity metric to capture the diversity-performance correlation among component LLMs of an ensemble. (ii) We develop a diversity-optimized ensemble pruning algorithm to select the top-k sub-ensembles from a pool of $N$ base LLMs. Our pruning method recommends top-performing LLM subensembles of size $S$, often much smaller than $N$. (iii) We generate new output for each prompt query by utilizing a learn-to-ensemble approach, which learns to detect and resolve the output inconsistency among all component LLMs of an ensemble. Extensive evaluation on four different benchmarks shows good performance gain over the best LLM ensemble methods: (i) In constrained solution set problems, LLM-TOPLA outperforms the best-performing ensemble (Mixtral) by 2.2\% in accuracy on MMLU and the best-performing LLM ensemble (MoreAgent) on GSM8k by 2.1\%. 
(ii) In generative tasks, LLM-TOPLA outperforms the top-2 performers (Llama70b/Mixtral) on SearchQA by $3.9\mathrm{x}$ in F1, and on XSum by more than $38$ in ROUGE-1. Our code and dataset, which contains outputs of 8 modern LLMs on 4 benchmarks is available at \url{https://github.com/git-disl/llm-topla} 
\end{abstract}

\section{Introduction}
Modern Large Language Models \citep{achiam2023gpt, jiang2024mixtral, touvron2023llama, team2024gemma} are characterized by architectures with billions of parameters, massive training datasets, and remarkable performance across many zero and one-shot tasks.  
Recently, there has been a myriad of open-sourced models, aiming for improving generalizability in a subset of tasks e.g., question answering, code generation, multi-agent, and summarization, with smaller sizes ($1$b to $70$b), and yet performing equally or better compared to larger sizes on that particular task \cite{zhao2023survey, hassid2024larger, mei2024bigger, hu2024survey}. This enables LLM consumers to access many open-source LLMs of various sizes and choose to run them locally or via an API from an LLM inference service provider.
A widely recognized challenge is how to select among the large collection of open/close-sourced LLMs the best model combination, and how to combine possibly conflicting output answers from multiple LLMs to reach the best generative output for the target learning task.

We argue that a practical LLM ensemble method should provide an efficient solution approach to answer both of the above questions. To this end, first, we introduce LLM-TOPLA, a diversity-optimized LLM ensemble method with three unique properties: (i) a focal diversity metric to capture the error diversity and the diversity-performance correlation among component LLMs of an ensemble; (ii) a diversity-optimized ensemble pruning algorithm to identify and select the top-k sub-ensembles from a pool of $N$ base LLMs, which shows equal or better performance compared to the ensemble of $N$ models; (iii) a learn-to-combine approach, which learns to detect and resolve the output inconsistency among all component LLMs of an ensemble, and generate the LLM-TOPLA output for each prompt query.

\section{Related Work} 
We broadly categorize the related work in achieving better generalization performance of LLMs into two threads: ensemble with unsupervised or supervised learning. 

In unsupervised methods, prompt engineering, exemplified by Chain of Thought (CoT) \citep{wang2022self}, generates multiple solution passes, with majority voting used to ensemble the final output. The downside of majority voting is the definition of equality between divergent answers. Compared to math problems or multiple-choice problems, consensus-based approaches like weighted majority voting may do poorly for generative queries.
Recently, two threads of research to further improve CoT. One advocates integrating more agents (models) from different LLM producers~\cite{li2024more} and utilizing the BLEU score as the heuristic to compare answers. Another is to enhance the BLEU score-based answer combination method by either assigning weights \citep{yao2024tree} or by creating a debate environment \citep{wan2024knowledge}. One caveat in common for these unsupervised methods is that they require lengthy and complex prompt strategies. 

Several supervised LLM ensemble methods are proposed: LLM-Blender~\citep{jiang2023llm} performing two steps of training; one for model selection and one for generation. Yet, the proposed ranking model requires pairwise comparison of models in the pool and the ensemble method has a limited context window with the high cost of training. Alternatively, a distillation strategy is proposed in \citep{wan2024knowledge} by performing a token alignment on the probability distributions of the models. In addition to the high computational cost, this paper only ensembles LLama-2 architectures. Regarding the model selection, \citep{chen2023frugalgpt} reduced the cost of inference by performing prompt adaptation, caching, and model tuning to choose the strongest model in the pool. We extensively evaluate on multiple-choice, open-ended, and generative question benchmarks to show that the proposed LLM-TOPLA outperforms the best LLM ensemble methods on MMLU, GSM8k, SearchQA and XSum. 


\begin{figure*}[t]
\centering
    \begin{subfigure}{0.33\textwidth}
        \centering
        \includegraphics[width=\textwidth]{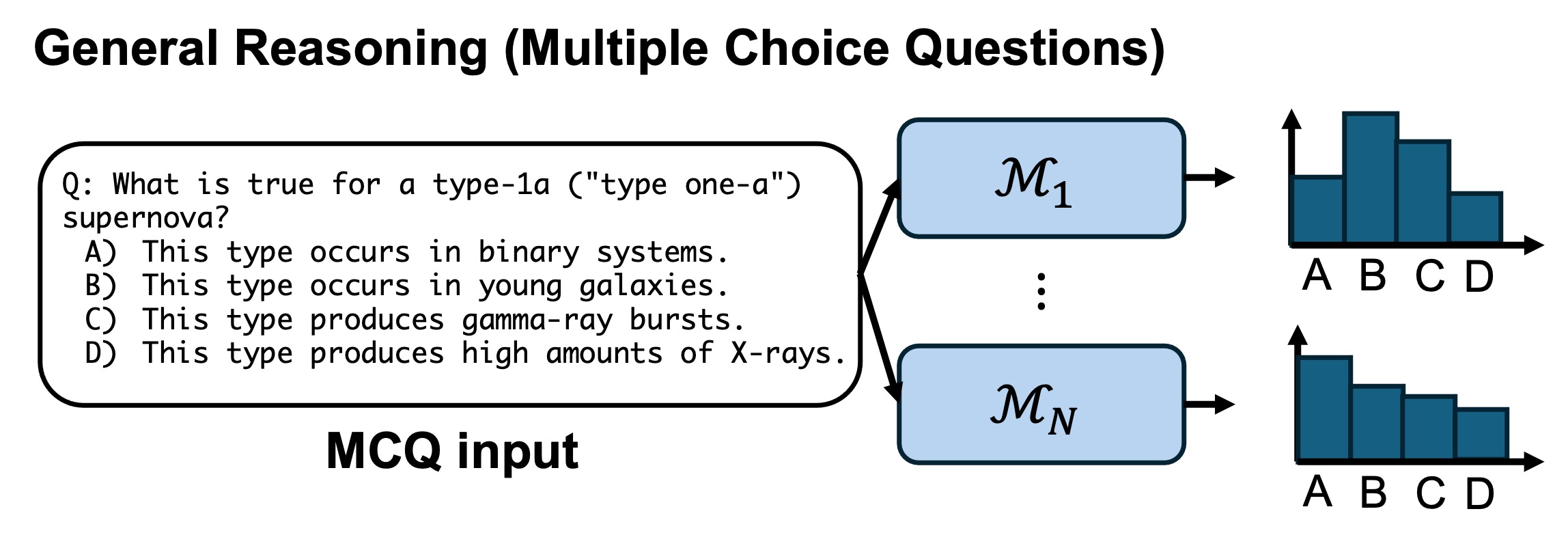}
        \vspace{-15pt}
        \label{fig:mcq}
    \end{subfigure}
    \begin{subfigure}{0.33\textwidth}
        \centering
        \includegraphics[width=\textwidth]{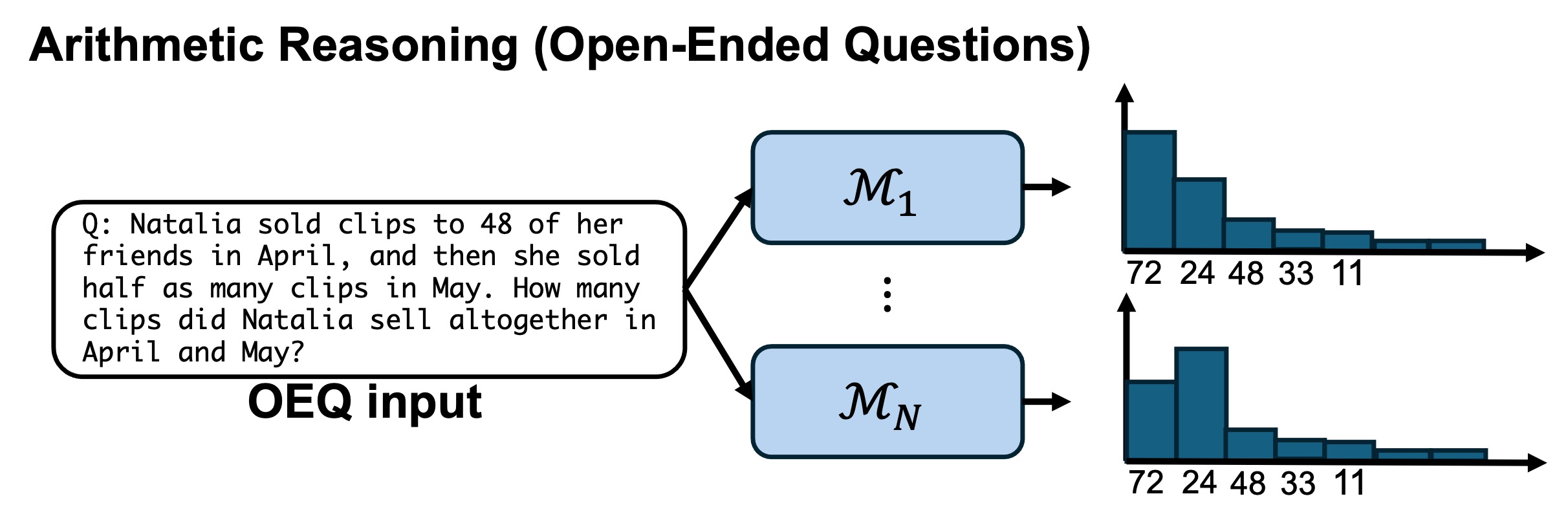}
        \vspace{-15pt}
        \label{fig:oeq}
    \end{subfigure}
    \begin{subfigure}{0.32\textwidth}
        \centering
        \includegraphics[width=\textwidth]{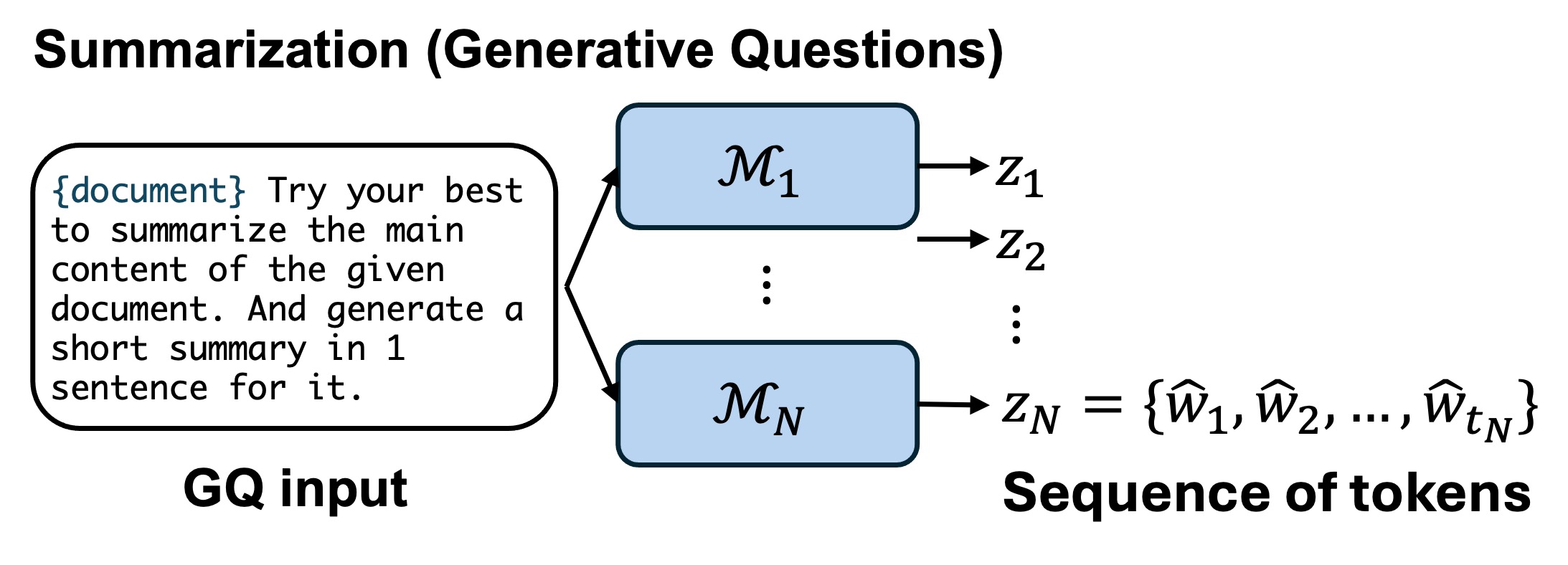}
        \vspace{-15pt}
        \label{fig:gq}
    \end{subfigure}
    \caption{We present the different types of tasks with their solution spaces.}
    \label{fig:scatter}
    \vspace{-8pt}
\end{figure*}

\section{Problem Definition}
Let $x$ denote an input query for task $T$ under an LLM $\mathcal{M}$ and $y$ represent the desired output. We assume a dataset $\mathcal{D}$ to be the collection of samples for task $T$, such that $(x, y) \in \mathcal{D}$. For a pool of LLMs with the size $N$, denoted as $\mathcal{M}_1, \dots, \mathcal{M}_N$, we utilize $\mathcal{D}$ to find the optimal ensemble function. This function takes outputs of each LLM and yields one final answer, denoted as $\tilde{y}$, given by $f(\mathcal{M}_{1}(x), \dots, \mathcal{M}_{N}(x))=\tilde{y}$, such that the difference between desired output is minimized, measured by the loss function $\mathcal{L}(\tilde{y}, y)$. However, based on the task $T$, the desired output $y$ can represent different solution spaces. Here, we define three different types of solution spaces.

In the first type, $y^{(1)} \in \{1, \dots, m\}$ represents the choices in a multiple-choice question (MCQ), where $m$ is typically a small integer, such as 4. The second type of outputs represents the type of open-ended question (OEQ), such as an answer to a multi-step descriptive math problem where the expected answer is a real number and denoted as $y^{(2)} \in \mathbb{R}$ or the expected answer can be a word representing the short answer to a trivia question denoted as $y^{(2)} \in \{w_1, \dots, w_{|V|}\}$, where $|V|$ is the size of the vocabulary. Lastly, the third type represents the outputs of generative question (GQ) tasks such as machine translation, summarization, and open question-answering. The solution space consists of a sequence of words, given by $y^{(3)}=\{w_1, \dots, w_{t}\}$. As illustrated in Figure \ref{fig:scatter}, the key difference between other solution sets is that the third solution type is a sequence of words whereas the second solution set consists of an exact word or a number. Next, we describe how our proposed methodology addresses all the solution sets by introducing ensemble learning functions for each type of problem.

\section{Ensemble Learning Functions}
We propose two learn-to-ensemble methods. The first method, {\sc TOPLA-Weighted}, is lightweight and applicable only to the first two types of outputs $y^{(1)}$ and $y^{(2)}$. The second method, {\sc TOPLA-Summary}, applies to all three types at a higher cost of complexity.

\subsection{LLM-TOPLA-Weighted}
An autoregressive language model predicts the next token, $w_{t}$, based on the probability mass conditioned on the input query, $x$, and the formerly generated tokens, $w_{<t}$, i.e. it models: 
\begin{equation}
    p(w_{t}|x, w_{<t}) = \frac{\exp(c_{t-1})}{\sum_{j=1}^{|V|} \exp(c_j)},
    \label{next_word_prob}
\end{equation}
where $c$ represents the output vector of the final linear layer of a language model. For an MC question, as proposed in \citep{hendrycks2020measuring}, the probabilities assigned to choices are obtained by calculating the probability of the choice's token using equation $\ref{next_word_prob}$. For instance, $p(w_t=\mathrm{A}|x, w_{<t})$ is calculated for choice A. However, a more popular methodology proposed by \citep{eval-harness} is used by the HuggingFace Leader Board \cite{open-llm-leaderboard} and also in our paper. We aggregate the probabilities of the tokens creating the whole choice to compute the probability of an answer. After repeating the procedure for all the choices, we obtain the probability distribution over the choices, denoted by $\mathbf{q}=[q_1, \dots, q_m]$, where $q$ represents the probability of a choice and $m$ is the number of choices.

As shown by\citep{holtzman2021surface}, the highest probability answer may not lead to a correct decision, and the probabilities assigned to other choices carry equal significance. Furthermore, we defend that the probability distribution of a model defines its characteristics and multiple models can be leveraged to reach the correct answer. To this end, we aim for the most robust way to combine $N$ different probability distributions, denoted by $\mathcal{M}_i(x)=\mathbf{q}_i$ where $i=1,\dots,N$ to generate the ensemble output, $\tilde{y}$, against the query $x$ sampled from a dataset $\mathcal{D}$. Our goal is to maximize the probability of the correct choice conditioned on the probabilities of base models:
\vspace{-4pt}
\begin{equation}
    \max \sum_{(x,y) \in \mathcal{D}} p(y|\mathcal{M}_{1}(x), \dots, \mathcal{M}_N(x)).
    \vspace{-8pt}
\end{equation}
We approximate this likelihood using an ensemble learner parameterized by $\theta$:
$
    f(\mathbf{q}_1, \dots, \mathbf{q}_N; \theta) = \tilde{y}.
$
This ensemble learner can be a machine learning model such as decision trees, or a neural network. In this paper, we use a Multi-layer Perceptron (MLP) containing multiple layers of fully connected weights with sigmoid activation functions. At the final layer, the model performs softmax to produce the output probability:
\vspace{-4pt}
\begin{equation}
    \tilde{y} = \mathrm{softmax}(\mathbf{W}_{H} (\dots \sigma(\mathbf{W}_{1}[\mathbf{q}_{1}, \dots,\mathbf{q}_{N}])\dots)),
    \label{eq:fusion}
\end{equation}
where $H$ is the number of layers. The first layer takes the concatenation of the probabilities as the input, i.e., $\mathbf{W}_1\in\mathbb{R}^{(mN)\times d}$ where $d$ is the input dimension of the second layer. We want to find the best parameters $\theta = (\mathbf{W}_{1},\dots,\mathbf{W}_{H})$ to maximize the likelihood, which can be reduced to minimize the cross-entropy loss on a dataset which is the collection of probabilities for each component model. Thus, we split the dataset into train, validation, and test and use the training set to train the ensemble model, the validation set to stop the training, and finally, we use the test set to calculate the performance. In each iteration of training, the parameters are updated by minimizing the loss function:
\begin{equation}
\begin{split}
    \theta_{\mathrm{best}} &= \argmin_{\theta} \sum_{x, y \in \mathcal{D}^{\mathrm{train}}} \mathcal{L}_{\mathrm{vote}}(y, \tilde{y}), \\
     \tilde{y} &= f(\mathcal{M}_{1}(x), \dots, \mathcal{M}_N(x); \theta), \\
    \mathcal{L}_{\mathrm{vote}}(y, \tilde{y}) &= -\sum_{i=1}^{m} y_i \log(\tilde{y}_i).
\end{split}
\end{equation}
We use SGD to perform updates on the parameters for every iteration. The ensemble learner analyzes
the probabilities assigned by each model and their confidence level. Thus, we train the learn-to-ensemble model to learn how to efficiently recognize the patterns among the predictions of each component model. This allows the ensemble learner to learn to make the correct choice even in the absence of consensus, instead of blindly relying on consensus voting algorithms, such as majority or plurality voting.

\textbf{Generalizing the formulation for $y^{(2)}$}: 
Considering the size of the solution set for $y^{(2)}$ can be large, concatenating probabilities for each token is impractical, especially since an answer may comprise a long sequence of tokens. It is essential to reduce the size of the solution set. 
Inspired by the Chain-of-Thought (CoT) prompting~\citep{wang2022self}, we consider two scenarios: 
(i) If a model is certain of its answer, multiple passes of the same query would result in the same reasoning paths with the same answers. (ii) When a model is uncertain, the decision is dispersed into multiple paths with different answers. Hence, we need a mechanism to find the correct output when the model is uncertain. 
To address both problems, we iterate the input query $K$ times with CoT prompting and count the occurrences of answers and divide by $K$ indicating the probability distribution of the model for that query. The answers sampled from a model create its solution set. For $N$ number of models, we can have at most $K\times N$ different answers. Let $Y_{j}=\{\hat{y}_{1}, \dots, \hat{y}_{K}\}$ represent the solution set of $j^{\mathrm{th}}$ model where $\hat{y}_{i}$ is the $i^{\mathrm{th}}$ answer of the model. We define a counting function to count the occurrence of an answer in the solution set denoted by $g(\hat{y}_i, Y_j) = \sum_{y\in Y_j}\mathbbm{1}(\hat{y}_i=y)$. However, each model can have its own solution set that is different than the others. By selecting the top-$K$ answers in all of the solution sets, we create one final solution set, denoted by $Y^{\mathrm{final}}$. Next, we compute the probability distribution for the final solution set generated by each model. This is done by dividing the frequency of each answer in the solution set of the model by the total number of passes, given by:
\begin{equation}
    \begin{split}
        \mathbf{q}_j &= [q_1, \dots, q_K],\; j=1,\dots,N \\
        q_i &= \frac{g(\hat{y}_i, Y_j)}{K},\; \hat{y}_i \in Y^{\mathrm{final}},
    \end{split}
\label{eq:multi_k}
\end{equation}

where $\mathbf{q}_j$ is the probability distribution of the $j^{\mathrm{th}}$ model on the solution set. By obtaining the probabilities, we use the same ensemble learner in equation \ref{eq:fusion} to learn the correct answer, leveraging the confidences of models for the input query.

\subsection{LLM-TOPLA-Summary}
We design the LLM-TOPLA learn-to-ensemble by summarization (LLM-TOPLA-Summary for short) with two objectives in mind. First, considering certain generative tasks, such as machine translation, and question-answering, the LLM-TOPLA-Weighted is not applicable without relaxing the definition of equality between different solutions. Even if the definition is relaxed by using comparison metrics such as BLEU score or distance metrics on the vector representation of the outputs, the use of TOPLA-weighted will select one of the answers generated by the best component model of the ensemble to create a TOPLA solution set. This may fail to produce the best generative output, even by utilizing heuristics on the relaxed definition. Second, our goal with TOPLA-Summary is to create an ensemble learner that applies to all types of tasks and generates its own output. 

LLM-TOPLA-summary performs learn to ensemble as follows. First, we employ another language model to generate a summary of the outputs produced by each model. Next, we use a sequence-to-sequence (seq2seq) model with encoder-decoder architecture~\citep{jiang2023llm} by concatenating the outputs of the component models of a chosen ensemble of $S$ base models with the input query and generating the final output of LLM-TOPLA. The fitness of the solution is limited by constraints such as context length, computation complexity, and training complexity. When the input length is short, it limits the number of models that can be fused and forces truncation on the outputs of component models. Also the self-attention mechanisms in encoder-decoder models have quadratic complexity \citep{beltagy2020longformer}. In response to these limitations, we implement sparse attention and global attention such that we can increase the context length up to 16396 tokens with 149 million parameters, and utilize a small training dataset.
Recall, $y^{(3)}=\{w_1, \dots, w_T\}$, where $T$ is the sequence length of the desired output. Each model in the pool generates the predicted sequence denoted by $\mathcal{M}_{i}(x)=\{\hat{w}_1, \dots, \hat{w}_{T_i}\}=z_i$ and $T_i$ is the sequence length of the $i^{\mathrm{th}}$ model output which can be different than $T$. Let $h$ be the seq2seq model with $\phi$ parameters, and $\mathcal{Z} = \{z_1, \dots, z_N\}$ be the collection of candidates. Our goal is to approximate the desired sequence probability conditioned on the input query and the model outputs, given by:
\begin{equation}
    p(y|x, \mathcal{Z}) \approx h(x, \mathcal{Z}; \phi).
    \vspace{-4pt}
\end{equation}
We give the input sequence, $x_s$, to the seq2seq model in the format of $x_s = \mathrm{concat}(x, z_1, \dots, z_N)$ and use special tokens as separators to indicate the beginning and end of the question or an answer. Consider an ensemble from 3 base models, the input below is sent to the TOPLA-summary model: 
\begin{equation}
\small
 \begin{split}
 x_s = \mathrm{<boq>} x \mathrm{<eoq>} \mathrm{<boc1>} z_1 \mathrm{<eoc1>} \\ 
 \mathrm{<boc2>} z_2 \mathrm{<eoc2>}\mathrm{<boc3>}z_3\mathrm{<eoc3>}. 
 \end{split}
\end{equation}
We use distinct tokens to indicate which model each candidate belongs to. As the number of models in an ensemble increases, the length of the input sequence to the seq2seq model, $\ell$, grows, resulting in a high computational cost in self-attention:
\begin{equation}
    \mathrm{Attention}(Q, K, V) = \mathrm{softmax}(\frac{QK^T}{\sqrt{d_k}})V
\end{equation}

where the operations are performed in each layer of the Transformer model architecture \citep{vaswani2017attention} and $Q$, $K$, and $V$ contain query, key, and value vectors for all the tokens. The result of the softmax function produces scores for each token. These scores are then multiplied by $V$, scoring each token in the input sentence against every other token. This process results in the complexity of $O(\ell^2 \times d)$, where $d$ is the embedding dimension.

To reduce the complexity and increase the context length, we employ the \textit{sliding window attention} pattern by \citep{beltagy2020longformer}. A fixed-sized window slides through tokens in each layer increasing the receptive field towards the top layers. For a window of size $a$, each token attends to the surrounding tokens within a range of $a/2$. 
This reduces the computation complexity to $O(\ell \times d \times a)$ which scales linearly with the input sequence.

Finally, the TOPLA ensemble learner evaluates the relation between the question and the answer given by each model to decide which answer suits the best. To stress the relation between the question and each candidate's answer, we employ \textit{selective global attention} on the tokens of $x$ of the input question. The global attention is the standard self-attention by scoring each token against every other token. With the sliding and global attention mechanism, we increase the context window length, reduce the computational complexity, and improve the performance.

Overall, LLM-TOPLA-summary is optimized by finding the best model parameter $\phi$ that will maximize the joint distribution over the target tokens $p(y|x, z_1, \dots, z_N;\phi)$. It performs auto-regressive generation using the following cross-entropy loss for a target summary $y=\{w_1,\dots, w_T\}$:
\begin{equation}
    \mathcal{L}_{sum} = -\sum_{t=1}^{T}\log p(w_t|w_{<t-1}, x, \mathcal{Z};\phi)
    \vspace{-4pt}
\end{equation}
We use SGD to perform updates on the parameters in each iteration. As the LLM-TOPLA-Summary model is trained, it learns to generate the correct token sequence by utilizing the information provided by each candidate answer and the TOPLA-summary evaluation results.

\section{Focal Diversity and Ensemble Pruning}
Given a pool of base LLMs as an ensemble, LLM-TOPLA first performs the focal diversity-based ensemble pruning for two reasons: 
First, the diversity among base models improves the ensemble performance \citep{breiman1996heuristics, dietterich2000ensemble}. Second, as we add more models to the ensemble pool, it becomes more expensive to prompt each model, and the input length of the ensemble model increases. Thus, the model selection for an ensemble set is essential. 
Consider a pool of $N$ base models, the total number of possible ensemble teams with size $S$ ($2\leq S \leq N$) is $2^{N}-N-1$~\citep{wu2021boosting}. 
A key question is how to perform ensemble pruning efficiently. We argue that the smaller ensemble size and the higher ensemble diversity, the better the generation performance of the ensemble. 
\vspace{4pt}

\textbf{Focal Negative Correlation \& Focal Diversity.\/}
The focal negative correlation metric, $\rho^{focal}$ is used to quantify the level of error diversity among the component models of an ensemble concerning each model within the ensemble. The focal diversity metric $\lambda^{focal}$ 
is used to quantify the general error diversity of the ensemble by taking into account all focal negative correlation scores of an ensemble. 
Let $\mathcal{E}^S$ denote an LLM ensemble composed of $S$ models: $\{\mathcal{M}_1, \dots, \mathcal{M}_S\}$, we choose one of the $S$ base models each time as the focal model to compute the focal negative correlation score of this ensemble, denoted as $\rho^{focal}(\mathcal{M}_i; \mathcal{E}^S)$. We define the focal diversity of this ensemble team by the average of the $S$ focal negative correlation scores.
The procedure of computing the focal negative correlation score of $\rho^{focal}$ is as follows: 
(i) select a base model among the set of $S$ base models as the \textit{focal} model, (ii) take all the validation episodes that the focal model has failed and calculate the focal negative correlation score, (iii) repeat the previous steps until all $S$ focal negative correlation scores are obtained. $\{\rho^{focal}_1, \dots, \rho^{focal}_S\}$, and (iv) compute the average over the scores to obtain the focal diversity of ensemble $\mathcal{E}^S$, denoted by $\lambda^{focal}(\mathcal{E}^S)$:
\begin{equation}
{\small
\begin{split}
\lambda^{focal}(\mathcal{E}^S)=\frac{1}{S}\sum_{\mathcal{M}_i \in \mathcal{E}^{S}} \rho^{focal}(\mathcal{M}_i; \mathcal{E}^S)\\
\rho^{focal}(\mathcal{M}_i; \mathcal{E}^S ) = 1 - \frac{P(2)}{P(1)}\\
P(2)=\sum_{j=1}^{S}\frac{j(j-1)}{S(S-1)}p_j, \;
P(1)=\sum_{j=1}^{S}\frac{j}{M}p_j
\end{split}
\vspace{-8pt}
}
\end{equation}

\begin{figure*}[hbt!]
    \centering
    \includegraphics[width=1\textwidth]{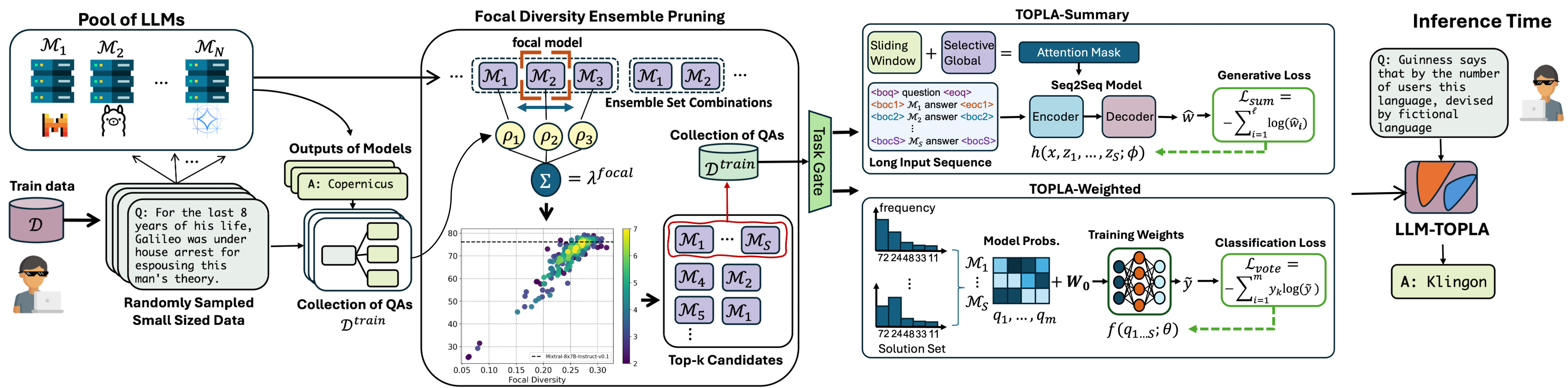}
    \caption{An overview of TOPLA-Framework.}
    \label{fig:ensemble}
\end{figure*}

Here $p_i$ is the probability that $i$ number of models fail together on a randomly chosen episode. We calculate as $p_i={n_i}/{L^{val}}$ where $n_i$ is the total number of episodes that $i$ number of models failed together on the validation set and $L^{val}$ is the total number of validation episodes. The term $P(2)$ represents the probability of two randomly chosen models simultaneously failing on an episode, while the denominator, $P(1)$, represents the probability of one randomly chosen model failing on an episode. The terms beneath $p_j$ values are the probability of the chosen model being one of the failures. For example, when $S=3$, there are three cases of model failures; one, two, or three models can fail simultaneously. If one model fails, the chance of selecting the failed model is $1/3$. Similarly, for two models, it is $2/3$, and for three models, it is $1$.
In the case of minimum diversity, the probability of two randomly chosen models failing together comes down to the probability of one of them failing, which makes the fraction term equal to 1 and $\rho^{focal} = 0$. Similarly, in the case of maximum diversity, there are no simultaneous failures. Hence, the nominator equals 0 and $\rho^{focal} = 1$. The definition of error changes according to the type of task and its solution set $y$. For the MCQs and OEQs, the errors are inequality between the prediction of the model and the label, for the GQs, the errors are missed 1-grams between the prediction and the label. Thus, the focal diversity captures member models that are not correlated solely by their error diversity.

\begin{figure}[hbt!]
    \centering
    \includegraphics[width=0.48\textwidth]{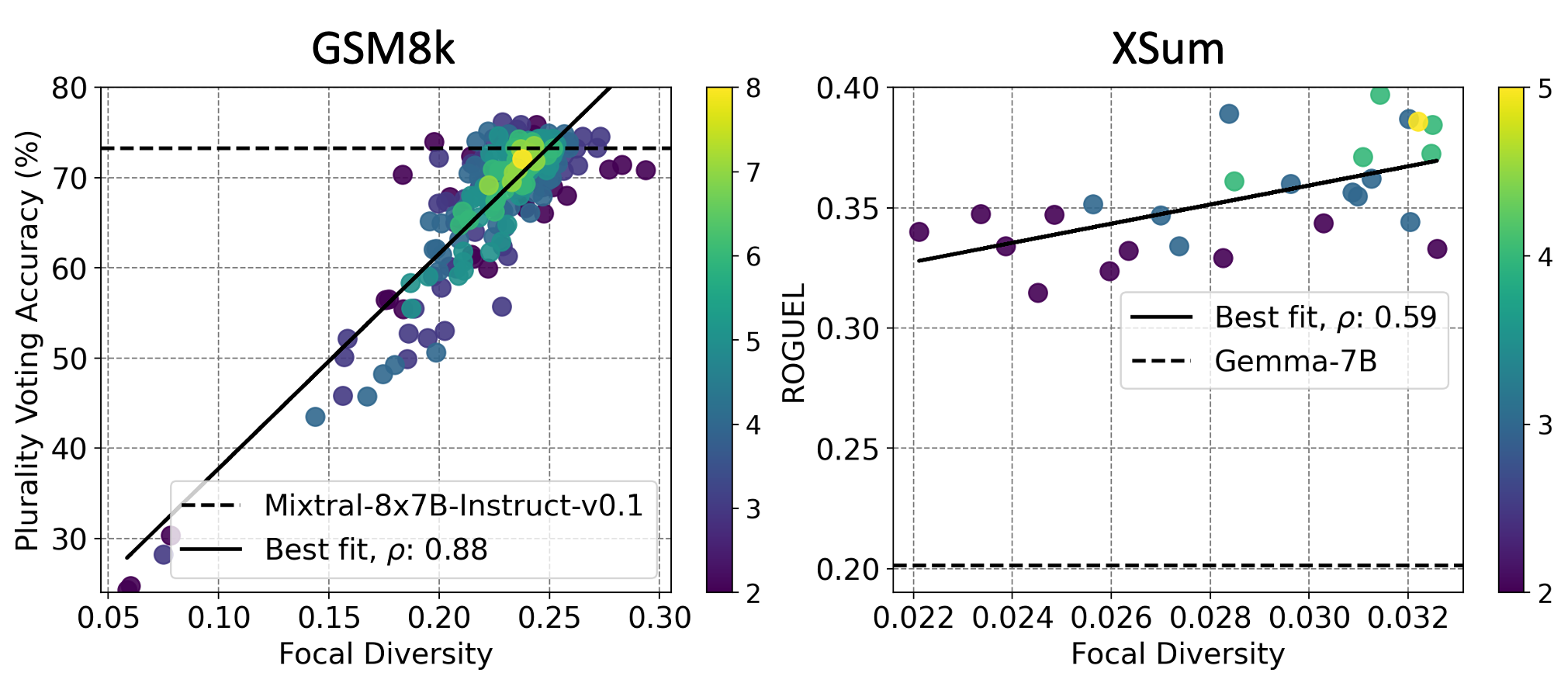}
    \vspace{-12pt}
    \caption{For each task, all candidate ensemble teams from the base model pools are plotted with their focal diversity scores and their performance metrics. The colors represent the size of each team, and the dotted line represents the best-performing individual model in the pool. We also plot the best-fit line with Pearson's Correlation Coefficient $\rho$ to show the correlation between performance and the focal diversity.}
    \label{fig:focal_first}
\end{figure}
\vspace{-2pt}  

\textbf{Ensemble Pruning Optimization.\/}
Figure~\ref{fig:focal_first} shows the focal diversity scores for a given pool of $N=8$ base models with GSM8k \citep{cobbe2021training} and XSum \citep{dunn2017searchqa} respectively ($N=6$ and see Appendix-B for the base models trained on other datasets). For GSM8k, we get $247$ candidate ensemble teams from the pool of $N=8$ base models. For XSum, we get $57$ candidate teams from the pool of $N=6$ base LLMs. 

We make three observations: (i) the focal-diversity metric is correlated with the model performance, (ii) there are multiple sub-ensemble teams of size 2-4 that outperform the largest ensemble of size $8$, and (iii) a majority of the smaller ensemble teams also outperform the best-performing individual model in the base model pool. 

To perform focal diversity-based ensemble pruning, we need to compute the focal diversity scores for all $2^{N}-N-1$ sub-ensemble teams when given a pool of $N$ base models. 
The brute force (BF) approach requires computing the focal diversity for each candidate ensemble of size $S$ ($2\leq S\leq N$). For $N=20$, we need to compute the focal diversity score for all $1,048,555$ candidate ensemble teams. 
To speed up this process, we leverage the Genetic Algorithm (GA) \cite{mirjalili2019genetic}, which takes significantly less time to reach the best combination. Table \ref{fig:speed} shows a comparison. For a pool of $N=15$ base models, we complete the focal diversity-based ensemble pruning in under a minute, achieving 5 orders of magnitude speed up (see Appendix C for further illustration and details).

\begin{table}[hbt!]
    \begin{adjustbox}{width=0.3\textwidth, center}
   \centering
    \small
    \begin{tabular}{p{1.1cm} p{1.1cm} p{0.7cm} p{0.8cm}}
      \hline
      \multirow{2}{1cm}{\# Base Models} & \multicolumn{3}{c}{Time(s)} \\
      \cline{2-4}
      & BF & GA & Gain\% \\
      \hline
      5 & 9.4 & 9.9 & -5.50 \\
      10 & 228.2 & 24.5 & 828 \\
      15 & 508.99 & 41.8 & 1116 \\
      \hline
    \end{tabular}
    \end{adjustbox}
\caption{Brute Force (BF) and GA pruning comparison.}
    \label{fig:speed}
  \vspace{-12pt}
\end{table}

\vspace{-2pt}
\section{LLM-TOPLA Framework}
\vspace{-4pt}
The framework for LLM-TOPLA is shown in Figure \ref{fig:ensemble}. The user inputs the training data, which includes queries with the desired outputs and a list of $N$ LLMs available in the pool. TOPLA will first send $\alpha \%$ of the queries to each LLM and generate $N$ outputs for each query. Each question and multiple answer pairs are stored to create $\mathcal{D}^{\mathrm{train}}$. In the second step, the focal diversity ensemble pruning module selects the best ensemble set combination to decrease pool size from $N$ to $S$ number of models, where $S<N$. It performs the Genetic Algorithm boosted diversity pruning algorithm and outputs top-k ensemble sets. Among top-k candidates, an ensemble set is selected randomly. Then, $\mathcal{D}^{\mathrm{train}}$ is updated based on the new $S$ model selection. At this point, the dataset can be populated with more samples if $\alpha < 100\%$, yet our experiments show that a small-sized dataset is enough to train a performant ensemble learner. According to the task type, TOPLA-framework will match the generated train data with either TOPLA-Summary or TOPLA-Weighted ensemble learner. After the training, the framework outputs LLM-TOPLA model which can be directly used by the user during inference time.

\section{Experiments}
\vspace{-4pt}
\begin{table*}[hbt!]
  \begin{adjustbox}{width=0.8\textwidth, center}
    \centering
    \small
    \begin{tabular}{l l l l c c c c}
        \hline
        \multirow{2}{1.7cm}{Model Name} & \multirow{2}{1.5cm}{Model ID} & \multicolumn{2}{c}{Inf. Time (s)$\downarrow$} & \multicolumn{2}{c}{MMLU\textsuperscript{*} test split (Acc \%)$\uparrow$} & \multicolumn{2}{c}{GSM8k\textsuperscript{\textdagger} (Acc \%) $\uparrow$} \\
        \cline{3-8}
        & & MMLU & GSM8k & HuggingFace LB & Together-AI & $K=1$ & $K=10$ \\
        \hline
        Phi-2 & 1 & - & 1.29 & $56.53_{0.91}$  & - & $51.09$ & $65.93$ \\
        Gemma-2b & 2 & 0.72 & 0.82 & $40.78_{0.57}$ & $31.41_{0.56}$ & $9.92$ & $19.56$ \\
        Gemma-7b & 3 & 1.44 & 0.87 & $65.26_{0.35}$ & $47.56_{0.53}$ & $53.50$ & $70.63$ \\
        Llama-7b & 4 & 4.82 & 1.58 & $42.62_{0.88}$ & $25.05_{0.60}$ & $8.08$ & $10.87$ \\
        Mistral-7b & 5 & 0.87 & 2.11 & $58.70_{0.86}$ & $40.04_{0.64}$ & $40.22$ & $54.02$ \\
        Llama-13b & 6 & 12.46 & 2.80  & $53.77_{0.53}$ & $44.40_{0.48}$ & $13.73$ & $19.02$ \\
        Llama-70b & 7 & 7.74 & 3.15 & $69.39_{0.96}$ & $51.60_{0.58}$ & $49.04$ & $56.52$ \\
        Mixtral-8x7b & 8 & 1.25 & 1.55 & $70.53_{0.95}$ & $64.82_{0.54}$ & $60.83$ & $71.16$ \\
        \hline
        LLM-TOPLA-Summary  & 378\textsuperscript{*}| 138 \textsuperscript{\textdagger} & 13.76 &  4.21 & - & $65.44_{0.96}$ & $65.40$ & $75.57$ \\
        LLM-TOPLA-Weighted  & 378\textsuperscript{*}| 138 \textsuperscript{\textdagger} & 12.46 &  4.05 & $\mathbf{72.77}_{1.18}$ & $\mathbf{65.75}_{0.93}$ & $\mathbf{66.82}$ & $\mathbf{79.01}$ \\
        \hline
    \end{tabular}
    \end{adjustbox}
    \caption{LLMTopla performance in MMLU and GSM8k dataset. We create the ensemble sets using focal-diversity on \textsuperscript{*} MMLU and \textsuperscript{\textdagger} GSM8k}
    \label{table:gsm8k}
\end{table*}

\begin{table*}[hbt!]
  \begin{adjustbox}{width=0.9\textwidth, center}
    \centering
    \small
    \begin{tabular}{l l c c c c c c c c }
        \hline
        \multirow{2}{1.7cm}{Model Name} & \multirow{2}{1.5cm}{Model ID} & \multicolumn{2}{c}{Inf. Time (s)$\downarrow$} &  \multicolumn{3}{c}{SearchQA\textsuperscript{*}} & \multicolumn{3}{c}{XSum\textsuperscript{\textdagger}} \\
        \cline{3-10}
        & & SearchQA & XSum & BLEU-1$\uparrow$ & EM (\%)$\uparrow$ & F1$\uparrow$ & ROUGE-1$\uparrow$ & ROUGE-2$\uparrow$ & ROUGE-L$\uparrow$ \\
        \hline
        Gemma-7b & 3 & 0.41 & 0.83 & $10.60$ & $4.43$ & $12.39$ & $26.43$ & $7.43$ & $20.13$ \\
        Mistral-7b & 5 & 0.36 & 1.59 & $4.12$ & $0.47$ & $5.15$ & $22.4$ & $5.49$ & $15.72$ \\
        Llama-13b & 6 & 0.39 & 1.97 & $8.77$ & $0.63$ & $10.6$ & $22.99$ & $6.25$ & $15.85$ \\
        Llama-70b & 7 & 0.32 & 1.64 & $13.97$ & $5.55$ & $15.95$ & $26.46$ & $7.70$ & $19.21$ \\
        Mixtral-8x7b & 8 & 0.38 & 1.21 & $13.13$ & $2.20$ & $16.04$ & $19.29$ & $5.47$ & $14.28$ \\
        \hline
        LLM-TOPLA-Summary  & 378\textsuperscript{*}| 3678 \textsuperscript{\textdagger} & 0.43 & 2.01 & $\mathbf{47.24}$ & $\mathbf{33.64}$ & $\mathbf{48.13}$ & $\mathbf{54.32}$ & $\mathbf{27.29}$ & $\mathbf{51.87}$\\
        \hline
    \end{tabular}
    \end{adjustbox}
    \caption{{LLMTopla performance in SearchQA and XSum dataset. We create the ensemble sets using focal-diversity on \textsuperscript{*} SearchQA and \textsuperscript{\textdagger} XSum}}
    \label{table:search_qa}
\end{table*}

We validate the effectiveness of LLM-TOPLA through extensive evaluations on MCQ, OEQ, and GQ benchmarks. We show that LLM-TOPLA outperforms the state-of-the-art LLM ensemble methods. Due to the space constraint, we include the details on the datasets and experimental setups in Appendix D.

\begin{table}[t]
    \begin{adjustbox}{width=0.5\textwidth, center}
    \centering
    \begin{tabular}{l c c c}
        \hline
        Method & Model ID & MMLU (Acc\%) {\textsuperscript{*}} & GSM8k (Acc\%) {\textsuperscript{\textdagger}}\\
        \hline
        More Agents & 6 & $51.09$ & $61.00$ \\
        More Agents & 7 & $60.05$ & $77.00$ \\
        LLM-Blender  & 12345678 & $44.01$ & $40.41$ \\
        Majority Voting  & 12345678 & $68.06$ & $72.31$ \\
        Mixtral-8x7b  & 8 & $70.53$ & $71.16$ \\
        \hline
        LLM-TOPLA & 378{\textsuperscript{*}}|138{\textsuperscript{\textdagger}} & $\mathbf{72.77}$ & $\mathbf{79.01}$ \\
        \hline
    \end{tabular}
    \end{adjustbox}
    \caption{ We compare our approach with the other ensemble methods in the literature.}
    \label{table:comparison_ens}
    \vspace{-8pt}
\end{table}

\begin{table}[t]
    \begin{adjustbox}{width=0.5\textwidth, center}
    \centering
    \begin{tabular}{l c c c }
        \hline
        \multirow{2}{1.7cm}{Method} & \multicolumn{3}{c}{XSum} \\
        \cline{2-4}
        & ROUGE-1$\uparrow$ & ROUGE-2$\uparrow$ & ROUGE-L$\uparrow$ \\
        \hline
        Pegasus 2B + SLiC & $49.77$ & $27.09$ & $42.08$ \\
        BRIO & $49.07$ & $25.59$ & $40.40$ \\
        LLM-TOPLA & $\mathbf{54.32}$ & $\mathbf{27.29}$ & $\mathbf{51.87}$ \\
        \hline
    \end{tabular}
    \end{adjustbox}
    \caption{ We compare our approach with previous SOTA methods of XSum, \citep{zhao2022calibrating} and \citep{liu2022brio}.}
    \label{table:comparison_xsum}
    \vspace{-8pt}
\end{table}

\subsection{Performance of LLM-TOPLA} 
\begin{figure*}[hbt!]
    \centering
    \includegraphics[width=1\textwidth]{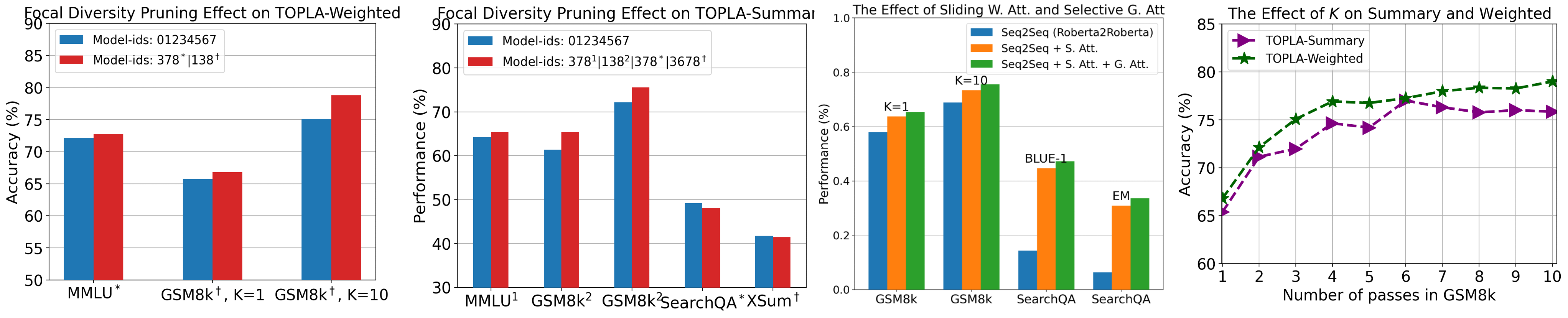}
    \caption{The effect of Focal-diversity Pruning is shown in the first two figures, and the effect of sliding window and selective global attention is shown in the third plot. Lastly, we show the effect of $K$ on TOPLA-Summary, and Weighted models in the GSM8k dataset.}
    \label{fig:graphs}
    \vspace{-8pt}
\end{figure*}

Table \ref{table:gsm8k} shows experiments on MMLU and GSM8k datasets, where we compare scores of each base model in the pool with the ensemble learners TOPLA-Weighted and Summary. The Model IDs of TOPLA denote the models in the ensemble set which is selected by focal diversity pruning. The inference time is the average response latency for a sample by each model in the pool. TOPLA framework sends each response in a parallel process, thus, the bottleneck is the slowest model, Llama-70b. In the MMLU dataset, TOPLA-Weighted reaches the best performance by surpassing the best-performing model Mixtral-8x7b by 2\%. Since the HF leader-board provides only the probability distribution of choices, we could only use the TOPLA-Weighted model for the predictions coming from HF. To test TOPLA-Summary we gather outputs from Together-AI API, however, the performance improvement on the best base model is marginal $<1\%$. We observe that the returned outputs do not change across multiple passes ($K>1$), thus preventing the ensemble model from considering alternative thoughts. In the GSM8k dataset, we provide scores when $K=1$ and $K=10$, where the outputs have high variation. While the TOPLA-Weighted model can improve the best-performing model by up to $6-8\%$, TOPLA-Summary improves $4-5\%$. As $K$ increases, the number of outputs leading to wrong thoughts rises, affecting the TOPLA-Summary model to reach the wrong conclusion, however, this effect is minimized by the frequency-based probability generation in the TOPLA-Weighted model. The full effect of $K$ on the performance is shown in Figure \ref{fig:graphs}.

Table \ref{table:search_qa} shows experiments on SearchQA and XSum datasets, where the TOPLA-Summary model ensembles the base models selected by focal diversity pruning. In the SearchQA dataset, TOPLA largely improves the best-performing model by up to $>30\%$ in the F1 score. When we look at the outputs generated by the models and the ensemble model as shown in Table \ref{table:example_search_qa} in Appendix F, we observe that the base models can gather related information about the question but the exact term is missing or either model is wordy and provides lots of unrelated information. TOPLA-Summary successfully detects the asked information gathered by each base model and generates the correct output. Each model has its expertise due to its training dataset's coverage and its learning capability. TOPLA can summarize and detect the asked information by exploiting the wisdom of models. Similarly, TOPLA-Summary surpasses the best-performing base model, Gemma-7b, by up to $>30\%$ in ROGUE-L score. By looking at the examples shown in Table \ref{table:example_xsum} in Appendix F, TOPLA-Summary provides a dense answer covering all the base model outputs and removing the redundancy. Using multiple base models allows the ensemble model to reach more-grained details on the sample document.

\subsection{Comparison with SOTA} 
Table \ref{table:comparison_ens} and Table \ref{table:comparison_time} (see Appendix E) presents a comparison of performance and time cost between LLM-TOPLA and other ensemble methods in the literature. More Agents \cite{li2024more} and LLM-Blender \cite{jiang2023llm} are the two well-known existing representative ensemble methods of pre-trained LLMs. We also add the majority voting method as a baseline to our approach.

More Agents is the previous SOTA and it uses a majority voting consensus method to combine multiple pre-trained LLMs, therefore, there is no training time. In open-ended questions, the authors adopt a BLEU score to find and select the closest answer with the highest accumulated similarity score. The approach is not generative and is bounded by the one of answers in the candidate set. Moreover, during inference, it must calculate the BLEU score for each pair in the candidate set of $n$ base models, the resulting complexity of $O(n^2)$. We showed that our LLM-TOPLA method outperforms More Agents on both closed QA and open-ended QA datasets. 

LLMBlender combines two supervised models: PairRanker, which ranks model outputs by pairwise comparison using a Deberta-v3-large (340M) backbone trained on 170k samples, and FLAN-T5-XL (3B) for output generation, trained on a large dataset. Using the top 3 candidates, LLMBlender concatenates outputs for final prediction. As shown in Table \ref{table:comparison_ens} our approach outperforms LLMBlender on both datasets. Additionally, our Focal Diversity pruning, which requires no training, ranks 10 models in 9.9 seconds, compared to PairRanker's 20.38-minute inference time and training on 170k samples.

Furthermore, to place TOPLA performance on XSUM dataset, we compare with the SOTA summarization methods in Table \ref{table:comparison_xsum}. LLM-TOPLA outperforms the SOTA on XSum dataset in all measures.

\subsection{Finetuning Base Models}
We also compare the performance of finetuned base models and TOPLA-Summary in the generative tasks, SearchQA and XSUM, using the same training data and a similar number of parameters. Therefore, we select FLAN-T5-Base \cite{flanx5} having 220 million parameters, Longformer encoder-decoder model (LED) \cite{beltagy2020longformer} with 149 million parameters, and Gemma-2 \cite{team2024gemma} containing 2 billion parameters with LoRA optimization the number of parameters significantly reduced to few million \cite{hu2021lora}. As shown in Table \ref{table:comparison_finetuned}, even though TOPLA uses less number of parameters, it outperforms finetuned versions of the base models. Particularly, Gemma-2 has a shorter training time ($2.35$ hours) compared to our approach ($2.41$ hours) but performs inferior compared to TOPLA. The cause is that the Gemma model is decoder-only, has a low context length, and LoRA reduces the number of trainable parameters.

\begin{table}[t]
    \begin{adjustbox}{width=0.5\textwidth, center}
    \centering
    \begin{tabular}{l c c c}
        \hline
        Method-finetuned & ROUGE-1 & ROUGE-2 & ROUGE-L \\
        \hline
        Gemma-2  & $15.5$ & $2.95$ & $11.98$ \\
        T5-large & $26.82$ & $6.94$ & $21.28$ \\
        LED  & $51.36$ & $24.05$ & $48.56$ \\
        TOPLA-Summary & $\mathbf{54.32}$ & $\mathbf{27.29}$ & $\mathbf{51.87}$ \\
        \hline
    \end{tabular}
    \end{adjustbox}
    \caption{We compare our approach with finetuned versions of LLMs on XSum dataset.}
    \label{table:comparison_finetuned}
    \vspace{-8pt}
\end{table}

\subsection{Experiments on BBH and ARC}
We expand our experiments on two challenging datasets, as shown in Table \ref{table:comparison_bbh}. Big-Bench Hard \cite{suzgun2022challenging} contains 23 challenging Big-Bench \cite{srivastava2022beyond} tasks, which language models have struggled to surpass compared to the average human rater. ARC \cite{arch} contains grade-school-level reasoning questions divided into two partitions: Easy and Challenge, with our focus on the latter. Both datasets are in multiple-choice (MC) format, and the base model predictions are obtained from the HF leader-board. As shown in Table \ref{table:comparison_bbh}, TOPLA-Weighted surpasses the best-performing base model in both datasets.

\begin{table}[t]
    \begin{adjustbox}{width=0.5\textwidth, center}
    \centering
    \begin{tabular}{l c c c}
        \hline
        Method & Model ID & BBH (Acc\%) & ARC (Acc\%) \\
        \hline
        phi-2b  & 1 & 44.55 & 56.29 \\
        gemma-7b  & 3 & 36.22 &	56.64 \\
        Mistral-7B  & 5 & 39.65 & 54.55 \\
        Llama-2-70b  & 7 & 28.02 & 48.95 \\
        Mixtral-8x7B & 8 & 44.12 & 63.29 \\
        \hline
        TOPLA-Weighted & 13578 & $\mathbf{ 	54.05}$ & $\mathbf{64.19}$ \\
        \hline
    \end{tabular}
    \end{adjustbox}
    \caption{The performance of base models and TOPLA-Weighted on BBH and ARC datasets. }
    \label{table:comparison_bbh}
    \vspace{-8pt}
\end{table}

\vspace{-4pt}
\subsection{Ablation Studies}
\vspace{-4pt}
To further observe the effect of the pruning and attention mechanisms, we execute two ablation studies in Figure \ref{fig:graphs}. First, we ensemble all the models in the pool and compare their performances with the ensemble model selected by the pruning mechanism. As shown in the first two figures, pruning improves the TOPLA-Weighted and -Summary in MMLU and GSM8k tasks and keeps the performance in SearchQA and XSum tasks. Although there is no improvement in the last two tasks, the pruned ensemble set is reaching the equivalent performance with fewer models. Second, we show the effect of the Seq2seq model, BART\citep{lewis2019bart}, sliding window attention, and selective global attention in the third figure by removing them in order and observing the resulting model performance in every dataset. In all of the tasks, all three combinations show the best performance.

\begin{figure*}[t]
    \centering
    \includegraphics[width=1\textwidth]{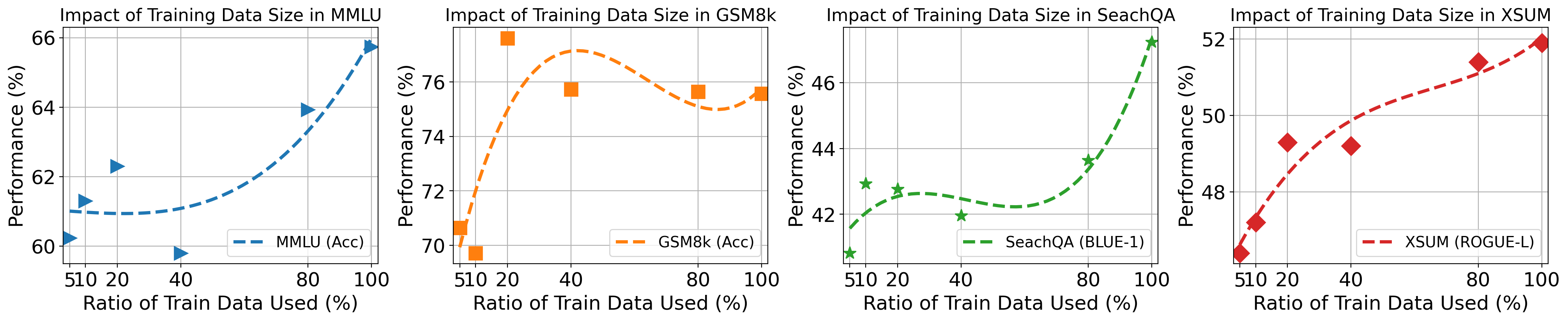}
    \caption{The effect of training data size to the performance.}
    \label{fig:data_size}
    \vspace{-8pt}
\end{figure*}

\vspace{-2pt}
\section{Conclusion}
\vspace{-4pt}
In this paper, we tackled the problem of ensembling modern LLMs from a wide perspective. The problem was defined as a mapping from three types of solution sets into the correct solution, and we introduced two different models. First, TOPLA-Weighted, the model attends weights to each base model output based on their confidence, and in the second type, we introduce a Seq2seq model, TOPLA-Summary, to perform summarization on concatenated outputs and generate one final answer. To stress the diversity, we created our ensemble set with the most diverse selection within seconds by Genetic Algorithm. The seq2seq model is further improved by employing sliding window attention to increase the context length and selective global attention to stress the relation between questions and answers. Our evaluation on 6 different benchmarks and 8 different modern LLMs shows that LLM-TOPLA framework outperforms the compared models and reaches SOTA.

Additionally, we provide a benchmark dataset that includes answers to MMLU, GSM8k, SearchQA, and Xsum, generated by the most popular large language models. This comprehensive dataset serves as a valuable resource for evaluating and comparing ensemble methods.

\textbf{Acknowledgments}
This research is partially sponsored by the NSF CISE grants 2302720, 2312758, and 2038029, an IBM faculty award, and a CISCO Edge AI program grant.

\section{Limitations}

The limitations of our study can be listed as the computational complexity and number of observational examples $\mathcal{D}^{\mathrm{train}}$. First, the main source of complexity is the cost of using multiple LLMs. In terms of user perspective, this burden is transferred to servers by the available inference services. The user can access each LLM with an API request. However, this aggregates the communication latency to the whole system. Therefore, we implement our framework in parallel so that the bottleneck is the slowest model. Second, we target the complexity of the pruning algorithm by employing the Genetic Algorithm, which allowed us to speed up the search by $>100\times$. Third, the complexity of the Seq2seq model is reduced by using a million-sized model, and we reduce the complexity coming from long input sequences by sliding window attention.

On the other hand, we assume an observational data $\mathcal{D}^{\mathrm{train}}$ which requires labeled samples. To investigate the effect of the training data size, we plot the effect of training data against performance in Figure \ref{fig:data_size}. The x-axis shows the percentage of training data we used from our dataset, e.g. in a total of 40,000 XSum samples and we used 5\% of them (8,000) to train and test it on the full portion of the test samples. The results demonstrate that even with a small ratio, the ensemble model enhances the performance of the best base model. However, as more data is used, the performance improves significantly. As a future direction, we will investigate the usage of synthetic data to decrease the dependency on labeled samples.


\bibliography{acl_latex}

\newpage
\appendix
\onecolumn

\section{Appendix}
\label{sec:appendix}
\section{Reproducibility Statement}
We make the following effort to enhance the reproducibility of our results. 
\begin{itemize}
    \item For {\sc LLM-TOPLA} implementation, a link to a downloadable source is included in our abstract. The link also includes the dataset of LLM outputs for each subtask.
    \item Our experiment details are given in Appendix D, containing selected hyperparameters.
    \item We also show the example outputs and prompts used in our paper in Appendix F.
\end{itemize}

\section{Speeding-up Ensemble Pruning with Genetic Algorithm}

The Genetic algorithm requires (i) the representation of a candidate solution, $\alpha$, and (ii) a fitness function, $r$, to evaluate the solutions. We represent each solution as a binary vector, where each index represents the presence of the base model in the ensemble set. For the fitness function, we create a focal pruning score metric on the validation dataset, by taking the convex combination of the focal diversity and other metrics such as the validation accuracy of each ensemble set (validation accuracy is applicable only for MCQ and OEQ, thus we used only focal-diversity score in GQ) or cost of models.

The pruning score calculation is given by, $r(\mathbf{\alpha}_i) = w_1\lambda_{i} + w_2 a_{i}$, where $a_i$ is the validation accuracy, $w_1$ and $w_2$ are the significance of each metric for pruning score such that $w_1 + w_2 = 1$ and $w_1, w_2 \in [0, 1]$. The initial population contains randomly created candidate solutions. During selection, the most fitted solutions survive to the next population. As the last step, we reproduce new solutions by performing a cross-over among the best-fitted solutions. The procedure is repeated until we reach a plateau or a predetermined fitness function value.

\begin{figure}[hbt!]
    \centering
    \includegraphics[width=0.5\textwidth]{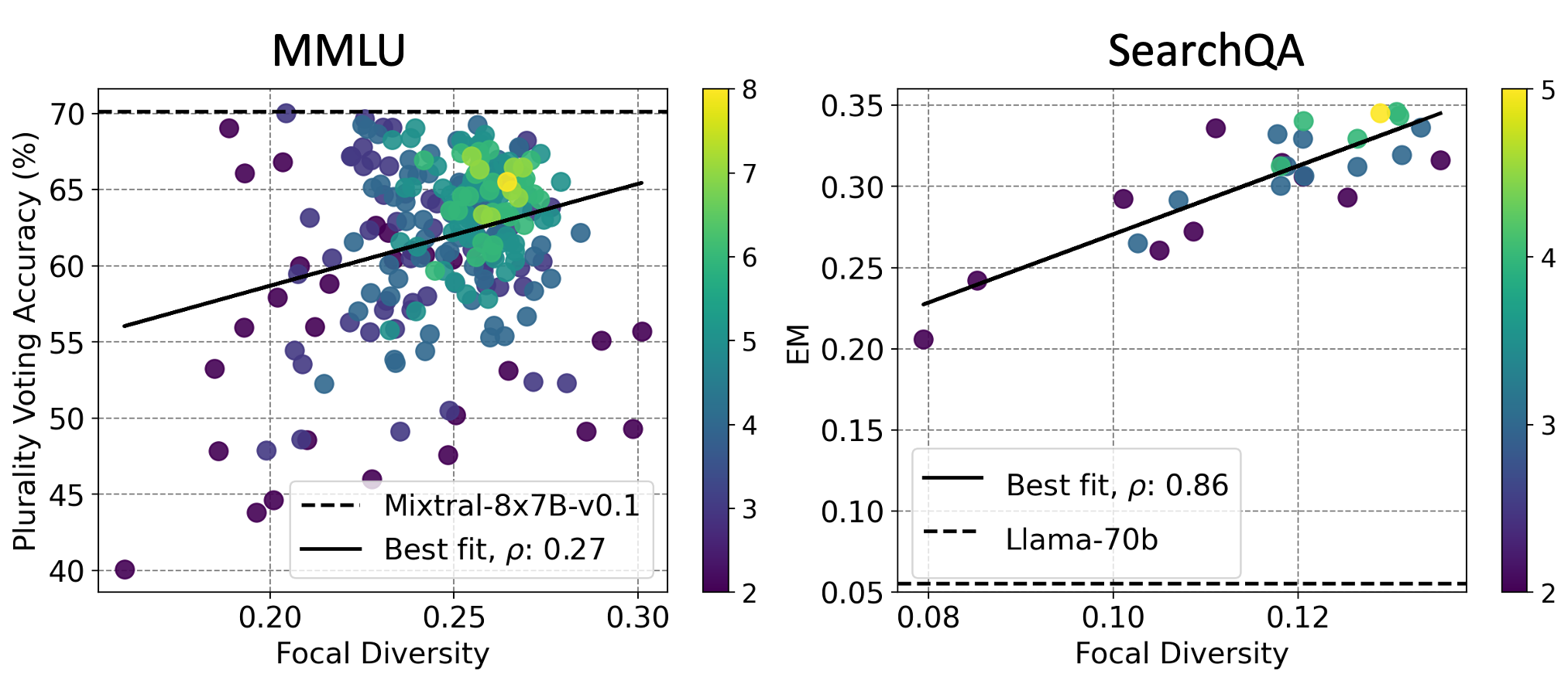}
    \caption{For MMLU and SearchQA tasks, we show all ensemble teams with their focal diversity scores and their performance metrics. The colors represent the size of each team, and the dotted line represents the best-performing individual model in the pool. We also plot the best-fit line with Pearson's Correlation Coefficient $\rho$ to show the correlation between performance and the focal diversity.}
    \label{fig:focal_1}
    \vspace{-8pt}
\end{figure}

\section{Datasets and Solution Spaces.} 
The experiments contain three different datasets targeting each type of solution space. For $y^{(1)}$, we use MMLU \citep{hendrycks2020measuring} which contains MCQs covering 57 subjects from STEM to social sciences with varying difficulties and total 14,042 samples. Our experiments on this dataset coming from two sources. The first is HuggingFace leader board \cite{open-llm-leaderboard} providing probability distribution of choices for each test sample. We also mimic a user who has only access to open-source LLMs through an API such as Together-AI \citep{togetherAI} or DeepInfra \citep{deepInfra}. However, currently, these APIs do not support next token probability distribution. Therefore, we performed regular expression parsing, plus, the highest BLUE-1 score between model output and the choices to find the output choice of the model. Performing multiple passes ($K>1$) allows us to obtain probability distribution for each sample, as we shown in equation \ref{eq:multi_k}. The dataset does not contain training samples; therefore, we perform a train-test split with a 70\% to 30\% ratio, repeat the procedure 20 times, and report the mean test score and standard deviation.

For $y^{(2)}$ type of solution spaces, we use GSM8k \citep{cobbe2021training}. The GSM8k dataset contains 7,472 training samples and 1,318 test samples, each with open-ended mathematical questions and multi-step solutions. Following \citep{wang2022self}, we perform CoT prompting on the base models up to $K=10$.

Lastly, for $y^{(3)}$, we measure the performance on generative tasks by employing SearchQA \citep{dunn2017searchqa} and XSum \citep{narayan2018don} datasets. The SearchQA dataset contains 172,908 train and 43,228 test samples where each sample is question-answer pairs with contexts. The questions are from\textit{Jeopardy!} and answers are 1-4 words length. We remove the contexts and performed closed-book prompting (see Appendix for examples and prompts). We used only 20,000 samples from train dataset to train our models and all the test samples to measure performance. On the other hand, the XSum contains 204,045 train and 11,334 test samples. Each sample includes a news article and one sentence summary. We used only 40,000 samples from train dataset to train our models and all the test samples to measure performance.

\textbf{Evaluation.} We use accuracy to evaluate MMLU and GSM8k datasets. In SearchQA, we use BLUE-1, Exact Match (EM), and F1 scores, while in XSum, we use ROGUE-(1, 2, L) scores for evaluation of models.

\textbf{Model Pool.} In our model pool selection, we aim for three elements and their effect on performance: (i) size of the model, (ii) model variety, and (iii) being open-source.

\textbf{Fusion Model.} LLM-TOPLA-Weighted model contains two fully-connected hidden layers with 100 neurons and sigmoid activations between the layers. The model weights starts from Xavier initialization and converges in 200 epochs optimized by Adam. To implement LLM-TOPLA-Summary, we employ Longformer-Encoder-Decoder \citep{beltagy2020longformer} model which is initialized from BART weights \cite{lewis2019bart}.

\textbf{Ensemble Pruning.} We selected $w_1=0.6$ and $w_2=0.4$ while scoring a candidate ensemble set to give more importance to the diversity in MMLU and GSM8k datasets. In SearchQA and XSum datasets, only focal diversity is used for pruning. The genetic algorithm stops when the fitness function does not change for 100 consecutive generations.

\section{Improvement by the Ensemble}
\begin{table}[hbt!]
    \begin{adjustbox}{width=0.5\textwidth, center}
    \centering
    \begin{tabular}{l c c c}
        \hline
        Ens-Method & Dataset & Train Time & Inference Time \\
        \hline
        More Agents  & XSUM & - & $0.021$ s \\
        LLM-Blender & XSUM & $2$ d & $19.1$ s \\
        TOPLA-Weighted  & GSM8k & $20.17$ s & $0.023$ s \\
        TOPLA-Summary & XSUM & $2.41$ h & $2.1$ s \\
        \hline
    \end{tabular}
    \end{adjustbox}
    \caption{The total time spent by each ensemble model.}
    \label{table:comparison_time}
    \vspace{-8pt}
\end{table}

\begin{table*}[hbt!]
    \begin{adjustbox}{width=1\textwidth, center}
    \centering
    \begin{tabular}{l c c c c c c c c}
        \hline
        \multirow{2}{1.5cm}{Method} & MMLU & GSM8k & \multicolumn{3}{c}{SearchQA} & \multicolumn{3}{c}{XSum} \\
        \cline{2-9}
        & Accuracy$\uparrow$ & Accuracy$\uparrow$ & BLEU-1$\uparrow$ & EM (\%)$\uparrow$ & F1$\uparrow$ & ROUGE-1$\uparrow$ & ROUGE-2$\uparrow$ & ROUGE-L$\uparrow$ \\
        \hline
        
        Longformer-Encoder-Decoder & $54.27$ & $3.34$ & $15.92$ & $4.04$ & $16.39$ & $51.36$ & $24.05$ & $48.56$ \\
        LLM-TOPLA-Summary & $\mathbf{65.44}$ & $\mathbf{75.57}$ & $\mathbf{47.24}$ & $\mathbf{33.64}$ & $\mathbf{48.13}$ & $\mathbf{54.32}$ & $\mathbf{27.29}$ & $\mathbf{51.87}$ \\
        \hline
    \end{tabular}
    \end{adjustbox}
    \caption{To analyze the improvement coming from the base models, we perform experiments on the Seq2Seq model, Longformer-Encoder-Decoder, by removing outputs of base LLM in the pool.}
    \label{table:comparison_led}
\end{table*}

Table \ref{table:comparison_led} shows the results of an experiment on the effect of base models on performance. In each task, we remove the outputs of each base LLM from the input sequence of the Seq2seq ensemble learner, Longformer-Encoder-Decoder model.  and fine-tune it purely by using the questions and their answers or documents and their summaries. As shown in the results, LLM-TOPLA outperforms in every task including summarization. This indicates that the ensemble learner can exploit the information gathered by the base LLMs and provide correct results.

\begin{table}[hbt!]
\begin{tabularx}{\textwidth}{X}
\hline
\textbf{Prompt}: "Answer very briefly by using at most 4 words." \\
\hline
\textbf{Question}: In devising the periodic table, Mendeleev laid out the elements in rows \& columns to show the different relationships when read each way; he's said to have been inspired by a version of this card game.\\
\textbf{Ground Truth:} solitaire\\
\hline
\textbf{Mixtral-8x7B:}  Solitaire, rearranging cards.\\
\textbf{Gemma-7b:} "Sure, here is the answer: The answer is "Uno".\\
\textbf{Llama-2-70b:} The periodic table was inspired by a version of the game Uno." \\
\textbf{LLM-TOPLA:} solitare \\
\hline
\end{tabularx}
\caption{A sample \textit{Jeopardy!} question taken from the SearchQA dataset is shown with the ground truth, each base LLM answer according to the prompt, and the answer of LLM-TOPLA.}
\label{table:example_search_qa}
\end{table}

\newpage
\begin{table*}[hbt!]
\begin{tabularx}{\textwidth}{X}
\hline
\textbf{Prompt}: "Try your best to summarize the main content of the given document. \\ And generate a short summary in 1 sentence only. Summary:" \\
\hline
\textbf{Document}:  "A judge had made the request because efforts to get them through "political channels" had failed, the lawyer said.
France, the former colonial power, has not yet commented on the request. Mr Sankara's widow and supporters have repeatedly accused France of masterminding his 1987 killing because he was a Marxist revolutionary. He was killed by soldiers in a hail of bullets shortly before a meeting of his cabinet in Burkina Faso's capital, Ouagadougou. Mr Sankara's assassination led to his second-in-command, Blaise Compaore, seizing power in a coup.
Mr Compaore steered Burkina Faso into a strong alliance with France, which retains close political, security and economic ties with many of its former colonies in Africa.
Burkina Faso opened an investigation into the killing after he was ousted from power in 2014.
Speaking at a press conference in Ouagadougou, the family lawyer, Benewende Stanislas Sankara, said an investigating judge had formally asked French authorities to declassify military documents to see whether France had played a role in the assassination.
A request had also been made to interview French officials who were involved in Burkina Faso's affairs at the time, the lawyer added.
Burkina Faso issued an arrest warrant for Mr Compaore in 2015, accusing him of involvement in Mr Sankara's killing.
He has repeatedly denied the allegation, but has refused to return to Burkina Faso to stand trial.
Mr Compaore is exiled in Ivory Coast, also a former French colony." \\
\hline
\textbf{Ground Truth:} "Burkina Faso has asked France to declassify documents documents related the assassination of former-President Thomas Sankara, a lawyer has the family has said. " \\
\hline
\textbf{Mixtral-8x7b:} " A lawyer for the family of Thomas Sankara, the former president of Burkina Faso who was assassinated in 1987, has asked French authorities to declassify military documents and interview French officials as part of an investigation into Sankara's death, accusing France of playing a role in the assassination."\\
\hline
\textbf{Gemma-7b:} **Summary:** Burkina Faso has requested that France declassify military documents and interview officials regarding the 1987 assassination of its former president, Thomas Sankara. " \\
\hline
\textbf{Llama-2-13b:} Sure! Here is a summary of the given document in one sentence:
A lawyer for the family of former Burkina Faso President Thomas Sankara has requested that French authorities declassify military documents to investigate whether France was involved in Sankara's 1987 assassination, which led to the coup by his successor Blaise Compaore, who has denied any involvement but is exiled in Ivory Coast."     Burkina Faso has requested that France declassify military documents and interview French officials regarding the 1987 assassination of the country's former president, Thomas Sankara, amid allegations that France was involved in the killing. "\\
\hline
\textbf{Llama-2-70b:} Burkina Faso has requested that France declassify military documents and interview French officials regarding the 1987 assassination of the country's former president, Thomas Sankara, amid allegations that France was involved in the killing. \\
\hline
\textbf{LLM-TOPLA-Summary:} Burkina Faso has asked France to declassify military documents about the killing of ex-president Thomas Sankara, a lawyer for his family has said.\\
\hline
\end{tabularx}
\caption{A sample document taken from XSum dataset is shown with the ground truth, each base LLM summarization according to the prompt, and the answer of LLM-TOPLA.}
\label{table:example_xsum}
\end{table*}

\section{Samples from SearchQA and XSUM}
Table \ref{table:example_search_qa} shows an example output by each model and TOPLA-summary for a question taken from the SearchQA dataset. While the majority of the models give incorrect answers, TOPLA agrees with the minority and produces the correct output.

Table \ref{table:example_xsum} shows an example summarization task performed by each base model and the TOPLA-summary. While every model produces the expected information in the ground truth, LLM-TOPLA produces the most dense and similar to the ground truth.

\end{document}